\documentclass{article}

\usepackage[preprint, nonatbib]{neurips_2025}
\usepackage{neurips_2025_mods}  

\newcommand{\mr}[2]{\multirow{#1}{*}{#2}}          
\newcommand{\mc}[2]{\multicolumn{#1}{c}{#2}}       
\newcommand{\eR}[2]{$\in \mathbb{R}^{#1 \times #2}$} 
\def\fline{\Xhline{2\arrayrulewidth}}              

\title{Slim attention: cut your context memory in half without loss -- \emph{K-cache is all you need for MHA}}

\author{Nils Graef\thanks{\texttt{info@openmachine.ai}}, \, Kayvan Zahiri, \, Andrew Wasielewski \\
  \href{https://openmachine.ai}{OpenMachine}}

\begin{document} \maketitle

\begin{abstract}
Slim attention shrinks the context memory size by 2x for transformer models with MHA (multi-head attention), which can speed up inference by up to 2x for large context windows. Slim attention is an exact, mathematically identical implementation of the standard attention mechanism and therefore doesn’t compromise model accuracy. In other words, slim attention losslessly compresses the context memory by a factor of 2. For encoder-decoder transformers, the context memory size can be reduced even further: For the Whisper models for example, slim attention reduces the context memory by 8x, which can speed up token generation by 5x for batch size 64 for example. And for the T5-11B model for example, the memory can be reduced by 32x because its MHA projection dimension is larger than $d_{\text{model}}$. See \citep{tricks} for code and more transformer tricks, and \citep{slim-video} for this paper's YouTube video.
\end{abstract}

\section{Calculate V from K}
\begin{figure}[h!] \centering  
  \includegraphics[scale=0.81]{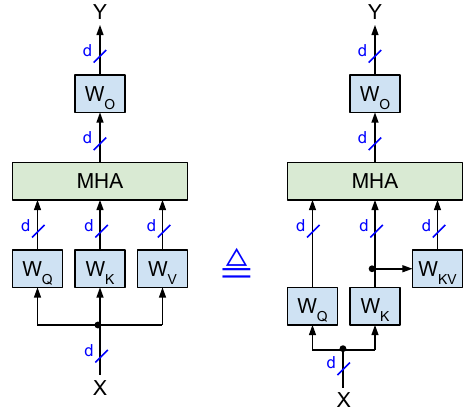}
  \caption{Mathematically identical implementations of multi-headed self-attention with square weight matrices \eR{d}{d}. \emph{Left}: vanilla version. \emph{Right}: proposed version where values V are computed from keys K with $W_{KV} = W_K^{-1} W_V$. The $\triangleq$ symbol denotes mathematical identity.}
\label{fig1} \end{figure}

Fig. \ref{fig1} illustrates how slim attention computes the value (V) projections from the key (K) projections in a mathematical equivalent way without hurting model accuracy. Therefore, we only need to store the keys in memory, instead of storing both keys and values. This reduces the size of the context memory (aka KV-cache) by half. Alternatively, slim attention can double the context window size without increasing context memory. However, calculating V from K on-the-fly requires additional compute, which we will discuss below.

Slim attention is applicable to transformers that use MHA (multi-head attention \citep{vanilla}) instead of MQA (multi-query attention \citep{MQA}) or GQA (grouped query attention \citep{GQA}), which includes LLMs such as CodeLlama-7B and Aya-23-35B, SLMs such as Phi-3-mini and SmolLM2-1.7B, VLMs (vision language models) such as LLAVA, audio-language models such as Qwen2-Audio-7B, and encoder-decoder transformer models such as Whisper \citep{whisper} and T5 \citep{T5}. Table \ref{tab1} lists various MHA transformer models ranging from 9 million to 35 billion parameters. The last column of Table \ref{tab1} specifies the KV-cache size (in number of activations) for each model to support its maximum context length, where the KV-cache size usually equals $2 d \cdot \text{layers} \cdot \text{context\_length}$.

\def\CodeLlamaSeven  {\href{https://huggingface.co/meta-llama/CodeLlama-7b-hf}          {CodeLlama-7B }}
\def\CodeLlamaThree  {\href{https://huggingface.co/meta-llama/CodeLlama-13b-hf}         {CodeLlama-13B }}
\def\CodeGemma       {\href{https://huggingface.co/google/codegemma-7b}                 {CodeGemma-7B }}
\def\ayaTwo          {\href{https://huggingface.co/CohereForAI/aya-23-35B}              {Aya-23-35B }}
\def\smollmTwo       {\href{https://huggingface.co/HuggingFaceTB/SmolLM2-1.7B}          {SmolLM2-1.7B }}
\def\smolvlm         {\href{https://huggingface.co/HuggingFaceTB/SmolVLM-Base}          {SmolVLM }}
\def\evoOne          {\href{https://huggingface.co/togethercomputer/evo-1-131k-base}    {Evo-1-131k }}
\def\PhiThree        {\href{https://huggingface.co/microsoft/Phi-3-mini-128k-instruct}  {Phi-3-mini-128k }}
\def\bitnet          {\href{https://huggingface.co/1bitLLM/bitnet_b1_58-3B}             {BitNet\_b1\_58-3B }}
\def\DCLM            {\href{https://huggingface.co/apple/DCLM-7B}                       {DCLM-7B }}
\def\olmoOne         {\href{https://huggingface.co/allenai/OLMo-1B-hf}                  {OLMo-1B }}
\def\olmoTwo         {\href{https://huggingface.co/allenai/OLMo-2-1124-13B}             {OLMo-2-1124-13B }}
\def\ChronosTiny     {\href{https://huggingface.co/amazon/chronos-bolt-tiny}            {Chronos-Bolt-tiny }}
\def\ChronosBase     {\href{https://huggingface.co/amazon/chronos-bolt-base}            {Chronos-Bolt-base }}
\def\Qwen            {\href{https://huggingface.co/Qwen/Qwen2-Audio-7B}                 {Qwen2-Audio-7B }}
\def\llavaNext       {\href{https://huggingface.co/llava-hf/LLaVA-NeXT-Video-7B-hf}     {LLaVA-NeXT-Video }}
\def\llavaVic        {\href{https://huggingface.co/llava-hf/llava-v1.6-vicuna-13b-hf}   {LLaVA-Vicuna-13B }}
\def\vicunaSeven     {\href{https://huggingface.co/lmsys/vicuna-7b-v1.5-16k}            {Vicuna-7B-16k }}
\def\vicunaThree     {\href{https://huggingface.co/lmsys/vicuna-13b-v1.5-16k}           {Vicuna-13B-16k }}
\def\flanBase        {\href{https://huggingface.co/google/flan-t5-base}                 {Flan-T5-base }}
\def\flanXXL         {\href{https://huggingface.co/google/flan-t5-xxl}                  {Flan-T5-XXL }}
\def\whisperTiny     {\href{https://huggingface.co/openai/whisper-tiny}                 {Whisper-tiny }}
\def\whisperLarge    {\href{https://huggingface.co/openai/whisper-large-v3}             {Whisper-large-v3 }}
\def\gpt             {\href{https://huggingface.co/openai-community/gpt2-xl}            {GPT-2 XL }}

\begin{table}[h!] \centering
\caption{\textbf{Various transformers with MHA} (instead of MQA or GQA) and their maximum KV-cache sizes (in number of activations) based on their respective maximum context length. $d$ is the embedding dimension (aka hidden size). See appendix for more MHA models.}
\begin{tabular}{lllccccc} \fline
  \thead[l]{Year} & \thead[l]{Publisher} & \thead[l]{Model} & \thead{Params} & \thead{$d$} & \thead{layers} & \thead{context \\ length} & \thead{context \\ memory} \\ \hline
  \mr{16}{2024} & \mr{2}{Meta}  & \CodeLlamaSeven \citep{code-llama}  & 7B    & 4,096         & 32              & 16k      & 4.3B  \\
                &               & \CodeLlamaThree \citep{code-llama}  & 13B   & 5,120         & 40              & 16k      & 6.7B  \\ \cline{2-2}
                & Google        & \CodeGemma \citep{codeGemma}        & 8.5B  & 3,072         & 28              & 8k       & 1.9B  \\ \cline{2-2}
                & Cohere        & \ayaTwo \citep{aya}                 & 35B   & 8,192         & 40              & 8k       & 5.4B  \\ \cline{2-2}
         & \mr{2}{HuggingFace}  & \smollmTwo \citep{smollm}           & 1.7B  & 2,048         & 24              & 8k       & 0.8B  \\
         &                      & \smolvlm \citep{smolvlm}            & 2.3B  & 2,048         & 24              & 16k      & 1.6B  \\ \cline{2-2}
         & Together AI          & \evoOne \citep{evo}                 & 6.5B  & 4,096         & 32              & 128k     & \textbf{34.4B} \\ \cline{2-2}
         & \mr{2}{Microsoft}    & \PhiThree \citep{phi3}              & 3.8B  & 3,072         & 32              & 128k     & \textbf{25.8B} \\
         &                      & \bitnet \citep{bitnet}              & 3.3B  & 3,200         & 26              & 2k       & 0.3B  \\ \cline{2-2}
         & Apple                & \DCLM \citep{dclm}                  & 6.9B  & 4,096         & 32              & 2k       & 0.5B  \\ \cline{2-2}
         & \mr{2}{Ai2}          & \olmoOne \citep{olmo}               & 1.3B  & 2,048         & 16              & 4k       & 0.3B  \\
         &                      & \olmoTwo \citep{olmo}               & 13.7B & 5,120         & 40              & 4k       & 1.7B  \\ \cline{2-2}
         & \mr{2}{Amazon}       & \ChronosTiny \citep{chronos}        & 9M    & 256           & 4               & 0.5k     & 1M    \\
         &                      & \ChronosBase \citep{chronos}        & 205M  & 768           & 12              & 0.5k     & 9.4M  \\ \cline{2-2}
         & Alibaba              & \Qwen   \citep{qwen2-audio}         & 8.4B  & 4,096         & 32              & 8k       & 2.1B  \\ \cline{2-2}
         & \mr{2}{LLaVA}        & \llavaNext \citep{llava-next}       & 7.1B  & 4,096         & 32              & 4k       & 1.1B  \\ \cline{1-1}
  \mr{3}{2023} &                & \llavaVic \citep{llava}             & 13.4B & 5,120         & 40              & 4k       & 1.7B  \\ \cline{2-2}
         & \mr{2}{LMSYS}        & \vicunaSeven \citep{vicuna}         & ~7B   & 4,096         & 32              & 16k      & 4.3B  \\
         &                      & \vicunaThree \citep{vicuna}         & ~13B  & 5,120         & 40              & 16k      & 6.7B  \\ \cline{1-2}
  \mr{4}{2022} & \mr{2}{Google} & \flanBase \citep{flan}              & 248M  & 768           & 12              & 0.5k     & 9.4M  \\
         &                      & \flanXXL  \citep{flan}              & 11.3B & 4,096         & 24              & 0.5k     & 101M  \\ \cline{2-2}
         & \mr{3}{OpenAI}       & \whisperTiny \citep{whisper}        & 38M   & 384           & 4               & 1500/448 & 6M    \\
         &                      & \whisperLarge \citep{whisper}       & 1.5B  & 1,280         & 32              & 1500/448 & 160M  \\ \cline{1-1}
  2019   &                      & \gpt \citep{gpt2}                   & 1.6B  & 1,600         & 48              & 1k       & 157M  \\ \fline
\end{tabular} \label{tab1} \end{table}
For long contexts, the KV-cache can be even larger than the parameter memory: For batch size 1 and 1 byte (FP8) per parameter and activation, the Phi-3-mini-128k model for example has a 3.8GB parameter memory and requires 25GB for its KV-cache to support a context length of 128K tokens. For a batch size of 16 for example, the KV-cache grows to 16 $\cdot$ 25GB = 400GB. Therefore, memory bandwidth and capacity become the bottleneck for supporting long context.

For a memory bound system with batch size 1, generating each token takes as long as reading all (activated) parameters and all KV-caches from memory. Therefore, slim attention can speed up the token generation by up to 2x for long contexts. For the Phi-3-min-128k model with 3.8GB parameters for example, slim attention reduces the KV-cache size from 25GB to 12.5GB, which reduces the total memory from 28.8GB to 16.3GB, and thus speeds up the token generation by up to 1.8x for batch size 1 (the maximum speedup happens for the generation of the very last token of the 128K tokens). And for batch size 16 for example, the speedup is (400+3.8) / (200+3.8) = 2x.

Transformer \citep{vanilla} defines the self-attention $Y$ of input $X$ with $h$ attention-heads as:
\begin{equation}
  Q = X W_Q = \text{concat} \left( Q_1, \ldots, Q_h \right) \qquad \quad K = X W_K = \text{concat} \left( K_1, \ldots, K_h \right) \label{eq4}
\end{equation}
\begin{equation}
  V = X W_V = \text{concat} \left( V_1, \ldots, V_h \right) \qquad \quad Y = \text{concat} \left( \text{head}_1, \ldots, \text{head}_h \right) W_O  \label{eq5}
\end{equation}
\begin{equation}
  \text{head}_i = \text{attention} \left( Q_i, K_i, V_i \right) = \text{softmax} \left( \frac{Q_i K_i^\top}{\sqrt{d_k}} \right) V_i \label{eq2}
\end{equation}

without the causal mask for simplicity, and with $W_Q = \text{concat}(W_{Q,1}, \ldots, W_{Q,h})$, $W_K = \text{concat}(W_{K,1}, \ldots, W_{K,h})$, $W_V = \text{concat}(W_{V,1}, \ldots, W_{V,h})$, and with the head dimension $d_k$. The matrices $Q, K, V, W_Q, W_K$, and $W_V$ are split into $h$ submatrices, one for each attention-head. Input $X$, output $Y$, queries $Q$, keys $K$, and values $V$ are \eR{n}{d}, where $n$ is the current sequence length (in tokens) and $d = d_{\text{model}}$ is the dimension of the embeddings.

For MHA, the weight matrices $W_K$ and $W_V$ are usually square matrices \eR{d}{d}, which allows us to calculate V from K as follows: Refactoring equation (\ref{eq4}) as $X = K W_K^{-1}$ lets us reconstruct $X$ from $K$, which we can then plug into equation (\ref{eq5}) to get
\begin{equation}
  V = K (W_K^{-1} W_V) = K W_{KV} \text{ and } V_i = K W_{KV,i} \text{ and } W_{KV} = \text{concat}(W_{KV,1}, \ldots, W_{KV,h})
\label{eq6} \end{equation}
and $W_{KV,i}$ \eR{d}{d_v}. Fig. \ref{fig1} illustrates the modified attention scheme that calculates V from K according to equation (\ref{eq6}). For inference, $W_{KV} = W_K^{-1} W_V$ can be precomputed offline and stored in the parameter file instead of $W_V$. This requires that $W_K$ is invertible (i.e. non-singular). In general, any square matrix can be inverted if its determinant is non-zero. It’s extremely unlikely that a large matrix has a determinant that is exactly 0.

Equation (\ref{eq6}) is exact, but in practice V is calculated at finite precision, and V can lose accuracy if $W_K$ is badly conditioned, which the determinant does not reveal. The V-cache scheme of appendix \ref{sec:vcache}, which calculates K from V, is an alternative. However, that scheme does not support RoPE. Section \ref{sec:experiments} compares both schemes in float32 for the decoder cross-attention of Whisper.

\textbf{Related work.} Slim attention is somewhat similar to DeepSeek’s multi-head latent attention (MLA) \citep{deepseek-v2}. Unlike MLA, slim attention is an exact post-training implementation of existing MHA models (including models with RoPE).

\section{K-cache is all you need}
\begin{figure}[h!] \centering
  \includegraphics[scale=0.71]{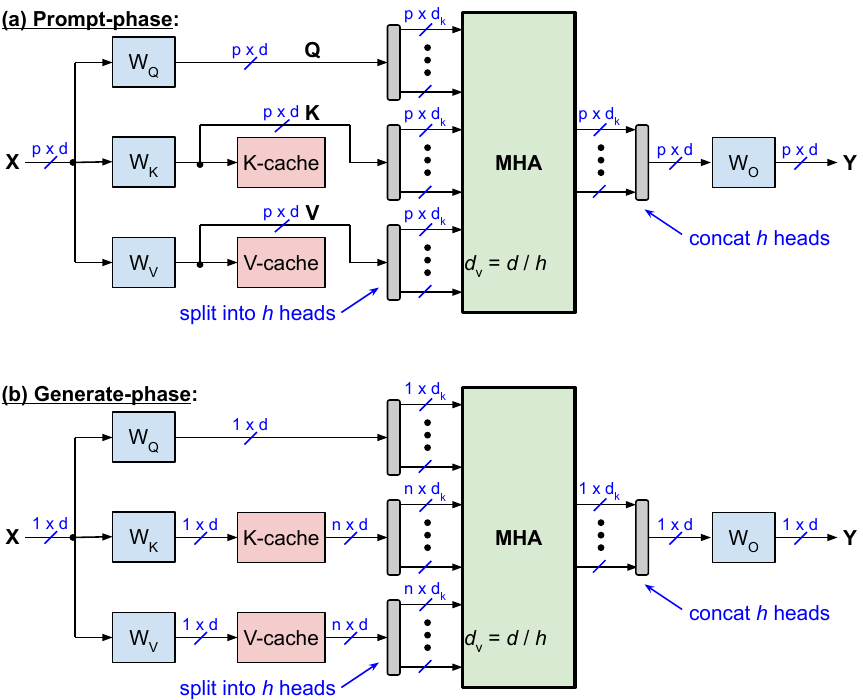}
  \caption{Standard MHA with KV-cache during (a) prompt-phase and (b) generate-phase.}
\label{fig2} \end{figure}

Inference consists of the following two phases, which are illustrated in Fig. \ref{fig2} for the vanilla MHA with KV-cache, where $p$ is the number of input-tokens and $n$ is the total number of current tokens including input-tokens and generated tokens, so $n = p+1, \ldots, n_{max}$ and $n_{max}$ is the context window length:
\begin{itemize}[topsep=-1pt]
  \item During the \textbf{prompt-phase} (aka prefill phase), all $p$ input-tokens are batched up and processed in parallel. In this phase, the K and V projections are stored in the KV-cache.
  \item During the \textbf{generate-phase} (aka decoding phase), each output-token is generated sequentially (aka autoregressively). For each iteration of the generate-phase, only one new K-vector and one new V-vector are calculated and stored in the KV-cache, while all the previously stored KV-vectors are read from the cache.
\end{itemize}

\begin{figure}[h!] \centering
  \includegraphics[scale=0.71]{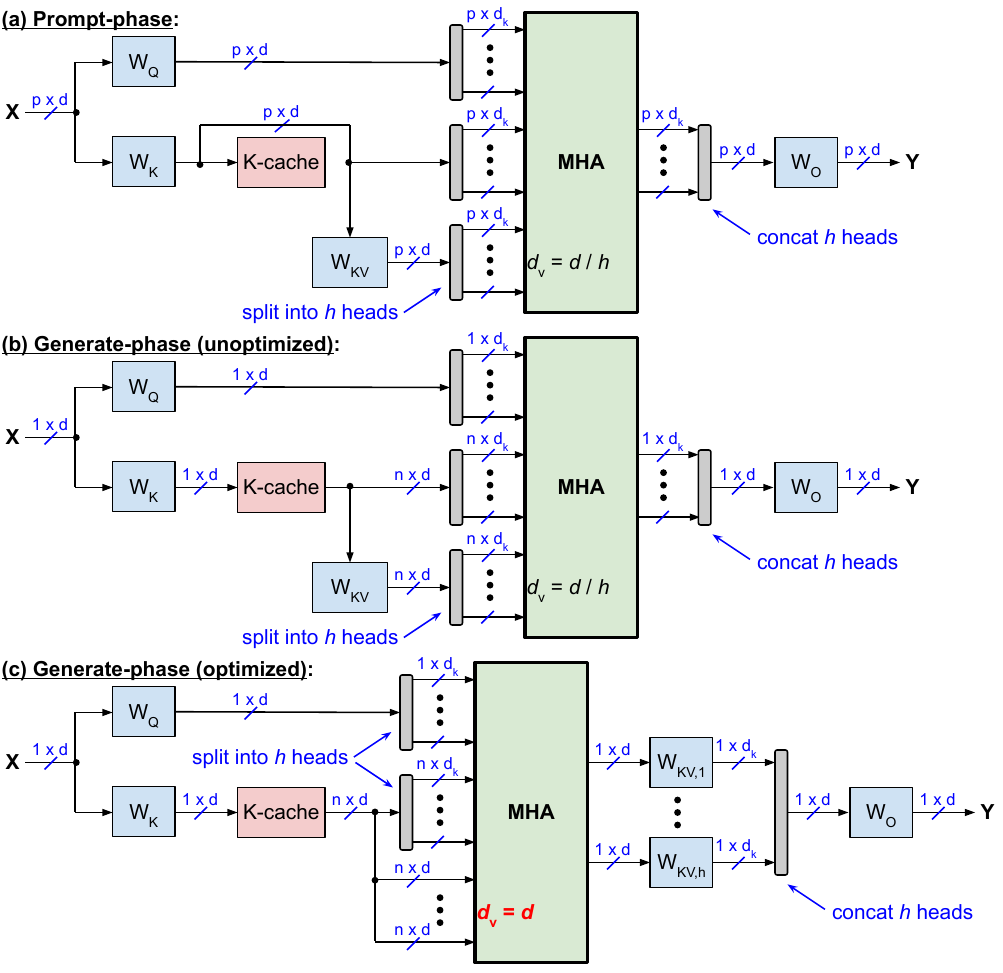}
  \caption{Slim attention without V-cache during (a) prompt-phase; (b) unoptimized and (c) optimized generate-phase.}
\label{fig3} \end{figure}

Fig. \ref{fig3} illustrates slim attention, which only has a K-cache because V is now calculated from K. Plugging equation (\ref{eq6}) into (\ref{eq2}) yields
\begin{equation}
  \text{head}_i
  = \underbrace{ \text{softmax} \left( \frac{Q_i K_i^\top}{\sqrt{d_k}} \right) \left[ K W_{KV,i} \right] }_{\text{Option 1}}
  = \underbrace{ \left[ \text{softmax} \left( \frac{Q_i K_i^\top}{\sqrt{d_k}} \right) K \right] W_{KV,i} }_{\text{Option 2}}
\label{eq7} \end{equation}
Equation (\ref{eq7}) can be computed in two different ways:

\begin{itemize}[topsep=-1pt]
  \item \textbf{Option 1 (unoptimized)}: Compute $V_i = K W_{KV,i}$ first, and then multiply it with $\text{softmax}(\cdot)$. This option is used by Fig. \ref{fig3}(a) and \ref{fig3}(b). Complexity: multiplying $K$ \eR{n}{d} with $W_{KV,i}$ \eR{d}{d_k} takes $2 n d d_k$ OPs\footnote{In general, multiplying two matrices with dimensions $m \times n$ and $n \times p$ takes $mnp$ MULs (two-operand multiply operations) and $mp(n-1)$ ADDs, so in total approximately $2mnp$ operations or OPs (two-operand operations).}, and multiplying $\text{softmax}(\cdot)$ \eR{1}{n} with the $n \times d_k$ result takes $2 n d_k$ OPs.
  \item \textbf{Option 2 (optimized)}: First multiply $\text{softmax}(\cdot)$ with $K$, and then multiply the result by $W_{KV,i}$.  This option is illustrated in Fig. \ref{fig3}(c). During the generate-phase, this option has lower compute complexity than option 1: multiplying $\text{softmax}(\cdot)$ with $K$ takes $2nd$ OPs, and multiplying the result with $W_{KV,i}$ takes $2dd_k$ OPs.
\end{itemize}

Option 2 above uses the same associativity trick as MLA, see appendix C of \citep{deepseek-v2}. During the prompt-phase, Fig. \ref{fig3}(a) has the same computational complexity as the vanilla scheme shown in Fig. \ref{fig2}(a). However, during the generate-phase, the proposed scheme has a slightly higher complexity than the vanilla scheme.

Table \ref{tab2} specifies the complexity per token per layer during the generate-phase for batch size 1. The columns labeled ``OPs'', ``reads'', and ``intensity'' specify the computational complexity (as number of OPs), the number of memory reads, and the arithmetic intensity, resp. We define the arithmetic intensity here as number of OPs per each activation or parameter read from memory. Specifically, the projection complexity includes calculating $X W_Q$, $X W_K$, $X W_V$, and multiplying with weight matrices $W_O$ and $W_{KV}$. And the memory reads for projections include reading all four weight matrices; while the memory reads of the MHA include reading the K-cache (and the V-cache for the vanilla implementation). See appendix for more details on MHA complexity.
\begin{table}[h!] \centering
\caption{\textbf{Complexity per token} per layer during the generate-phase for batch size 1}
\begin{tabular}{lcccccc} \fline
                                  & \mc{3}{Projection complexity}     & \mc{3}{MHA complexity}          \\
                                  & OPs         & reads   & intensity & OPs        & reads  & intensity \\ \hline
  Vanilla, see Fig. \ref{fig2}(b) & $8d^2$      & $4d^2$  & 2          & $4nd$      & $2nd$  & 2         \\
  Unoptimized, Fig. \ref{fig3}(b) & $(2n+6)d^2$ & $4d^2$  & $(n+3)/2$  & $4nd$      & $nd$   & 4         \\
  Optimized, Fig. \ref{fig3}(c)   & $8d^2$      & $4d^2$  & 2          & $2nd(h+1)$ & $nd$   & $2h+2$    \\ \fline
\end{tabular} \label{tab2} \end{table}

Note that for batch-size $B$, the arithmetic intensity of the vanilla transformer during the generate-phase is $2B$ for the FFNs and the attention-projections, but only 2 for the remaining attention operations (softmax arguments and weighted sum of V) because each of the $B$ tokens has its own KV-cache.
\begin{table}[h!] \centering
\caption{\textbf{TOPS\tablefootnote{tera-operations-per-second}, memory bandwidth, and arithmetic intensity} of popular chips}
\begin{tabular}{lccc} \fline
  Chip & \makecell{TOPS \\ (int8)} & \makecell{Theoretical memory \\ bandwidth (GB/s)} & \makecell{Arithmetic intensity \\ (OPs per byte)} \\ \hline
  Apple A18 \citep{apple-wiki}     & 35     & 60      & 583 \\
  Apple M4 Max \citep{apple-wiki}  & 38     & 410     & 93  \\
  Google TPU v4 \citep{TPU-wiki}   & 275    & 1,200   & 229 \\
  Google TPU v5p \citep{TPU-wiki}  & 918    & 2,765   & 332 \\
  NVIDIA H200 \citep{nvidia-wiki}  & 1,980  & 4,800   & 413 \\
  NVIDIA B200 \citep{nvidia-wiki}  & 4,500  & 8,000   & 563 \\ \fline
\end{tabular} \label{tab3} \end{table}

Table \ref{tab3} shows the arithmetic intensity (now defined as OPs per memory byte) of various SoCs, TPUs, and GPUs, which vary from 93 to 583. A system is memory bound (i.e. limited by memory bandwidth) if the arithmetic intensity of the executed program is below the chip’s arithmetic intensity. Here, the maximum arithmetic intensity of slim attention is $2h+2$, see Table \ref{tab2}, where $h$ is the number of attention-heads, which ranges between 16 and 64 for the models listed in Table \ref{tab1}. So the peak arithmetic intensity (up to 130 for $h = 64$) is usually less than the system’s intensity (except for Apple's M4 Max), which means that the system is still memory bound during the token generation phase. Therefore, slim attention speeds up the processing by up to 2x as it reduces the context memory reads by half. Furthermore, slim attention enables processing all heads in parallel as a single matrix-matrix multiplication instead of multiple vector-matrix multiplications, which is usually more efficient and faster on many machines. And slim attention is also compatible with Flash Attention \citep{flash-attention}, which performs softmax and value accumulation in parallel.

\section{Support for non-square KV weight matrices}
\begin{table}[h!] \centering
\caption{Models with non-square KV weight matrices where $e > d$}
\begin{tabular}{lccccc} \fline
  Model & $d$ & $d_k$ & $h$ & $e = d_k h$ & aspect ratio $r = e/d$ \\ \hline
  CodeGemma-7B  & 3,072  & 256  & 16   & 4,096   & 1.3 \\
  T5-3B         & 1,024  & 128  & 32   & 4,096   & 4   \\
  T5-11B        & 1,024  & 128  & 128  & 16,384  & 16  \\ \fline
\end{tabular} \label{tab4} \end{table}

Some transformers with MHA use non-square weight matrices for their K and V projections. Specifically, these models do not satisfy $d = d_k h$. Table \ref{tab4} shows three such models where $e = d_k h > d$. Let’s also define the aspect ratio $r$ as $r = e/d$. For example, Google’s T5-11B model has a large aspect ratio of $r = 16$. There are two options to reduce the KV-cache by 2x or more, which are compared in table \ref{tab5} and summarized as follows:
\begin{itemize}[topsep=-1pt]
  \item \textbf{Option 1}: Because the K weight matrix is non-square, inverting this matrix is not straight forward. And the resulting matrix $W_{KV}$ \eR{e}{e}, which has $r$-times more parameters than $W_V$ \eR{d}{e}.
  \item \textbf{Option 2}: Instead of storing  V in cache and then calculating V from K, we can store the smaller $d$-element vectors X before the projection and then on-the-fly calculate both projections (V and K) from X. The cache is now $r$-times smaller than option 1, and $2r$ times smaller than the baseline, for example 32 times smaller for the T5-11B model. However, this comes at a slightly higher computational cost.
\end{itemize}
\begin{table}[h!] \centering
\caption{Comparison of options for supporting non-square KV weight matrices}
\begin{tabular}{lccc} \fline
                            & Baseline & Option 1 & Option 2                         \\ \hline
  Cache reduction factor    & 1        & 2        & $2r$                             \\
  Size of $W_V$ or $W_{KV}$ & $d e$    & \makecell{$e^2$ ($r$-times larger)} & $d e$ \\
  Computational complexity  & baseline & higher   & even higher                      \\
  Support for RoPE?         & Yes      & Yes      & No                               \\ \fline
\end{tabular} \label{tab5} \end{table}

\textbf{Option 1}: The standard matrix inverse is defined only for square matrices, and the inversion functions in NumPy and SciPy are limited to such matrices. We want to compute the inverse of $W_K$ \eR{d}{e} with $e > d$ such that $W_K W_K^{-1} = I$, where $I$ is the identity matrix and $W_K^{-1}$ is the so-called right inverse of $W_K$. We compute $W_K^{-1}$ by using a trick that inverts the term $W_K W_K^\top$ instead of $W_K$ as follows:
\begin{equation*}
  I = W_K \underbrace{W_K^\top (W_K W_K^\top)^{-1}}_{W_K^{-1}}
\end{equation*}

In the equation above, everything on the right side of $W_K$ has to be the inverse of $W_K$, thus $W_K^{-1} = W_K^\top (W_K W_K^\top)^{-1}$. We can now use the matrix inversion function of NumPy to compute the inverse of the term $W_K W_K^\top$, which is a square $d \times d$ matrix. Now we can calculate $W_{KV} = W_K^{-1} W_V$. However, storing $W_{KV}$ instead of the original $W_V$ takes $r$ times more space in memory, which is an issue for large aspect-ratios $r$.

\begin{figure}[h!] \centering
  \includegraphics[scale=0.71]{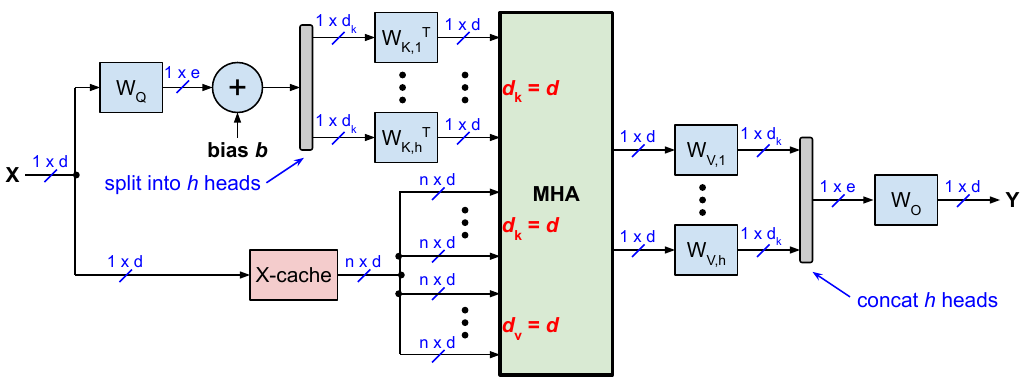}
  \caption{Slim attention with X-cache (instead of KV or V-cache) for the generate-phase of transformers with non-square weight matrices with $e > d$.}
\label{fig4} \end{figure}

\textbf{Option 2} caches the X-matrix instead of KV or just K, where the X-matrix contains the input activations of the attention layer (before the projections). Recomputing all K-vectors from X by multiplying X with weight matrix $W_K$ would require $2 n d e$ operations and would be very expensive. A lower complexity option is illustrated in Fig. \ref{fig4}, which is similar to the trick illustrated in Fig. \ref{fig3}(c).
Recall that for the $i$-th head ($i = 1, \ldots, h$), the softmax argument (without the scaling factor $1/\sqrt{d_k}$) is $A_i = Q_i K_i^\top$, where $Q_i = X W_{Q,i}$ and $K_i = X W_{K,i}$. For the generate-phase, there is only one input-vector $x_n$ for the query, but there are $n$ input-vectors $X$ for the key and value projections. We can take advantage of this and modify $A$ as follows (which uses the trick $(B C)^\top = C^\top B^\top$ for transposing the product of arbitrary matrices $B$ and $C$):
\begin{equation*}
  A_i = Q_i K_i^\top = x_n W_{Q,i} (X W_{K,i})^\top = (x_n W_{Q,i} W_{K,i}^\top) X^\top
\end{equation*}

For each iteration of the generate-phase, we now have to calculate the term $x_n W_{Q,i} W_{K,i}^\top$ only once for each attention-head, which is independent of the sequence length. Calculating this term involves multiplying the $d$-dimensional vector $x_n$ with matrices $W_{Q,i}$ \eR{d}{d_k} and $W_{K,i}^\top$ \eR{d_k}{d}, which requires $2 d e$ multiplications for the $h$ heads, so $4 d e$ operations in total (where we count a multiply-add operation as 2 operations).

This scheme also works for projection layers with biases (as used by the Whisper models for example). The appendix shows how we can eliminate the biases from the key and value projections, but not from the query projection. Adding a constant query bias-vector $b$ to the equation above is straightforward and also illustrated in Fig. \ref{fig4}:
\begin{equation*}
  A_i = Q_i K_i^\top = (x_n W_{Q,i} + b) (X W_{K,i})^\top = ((x_n W_{Q,i} + b) W_{K,i}^\top) X^\top
\end{equation*}

However, this scheme doesn’t work if there is a positional encoding such as RoPE located between the projection layers and the dot-product calculation. But option 2 fully supports other relative position encoding (PE) schemes such as RPE of the T5 model, Alibi, Kerple and FIRE \citep{FIRE} which add a variable bias to the softmax arguments (instead of modifying the queries and keys before the dot-product calculation). See for example the FIRE paper \citep{FIRE}, which shows that FIRE and even NoPE can outperform RoPE for long context.

\section{Support for encoder-decoder transformers} \label{sec:enc-dec}
In general, calculating K from V is not only possible for self-attention (see Fig. \ref{fig1}) but also for cross-attention. In this section, we present two context memory options for encoder-decoder transformers such as Whisper (speech-to-text), language translation models such as Google’s T5, and time series forecasting models such as Amazon’s Chronos models. One option is not limited to MHA only, but is also applicable to MQA and GQA. Table \ref{tab6} compares the options, which are summarized as follows:
\begin{itemize}[topsep=-1pt]
  \item The \textbf{baseline} implementation uses complete KV-caches for both self-attention and cross-attention of the decoder stack, which we refer to as self KV-cache and cross KV-cache, resp.
  \item \textbf{Option 1} is an optimized implementation where the V-caches for both self-attention and cross-attention are eliminated, which reduces the total cache size by 2x.
  \item \textbf{Option 2} eliminates the entire cross KV-cache and also eliminates the V-cache of the self-attention.
\end{itemize}
\begin{table}[h!] \centering
  \caption{Comparison of options for supporting encoder-decoder transformers}
  \begin{tabular}{lccc} \fline
                             & Baseline & Option 1 & Option 2 \\ \hline
  Self KV-cache size         & 100\%    & 50\%     & 50\%     \\
  Cross KV-cache size        & 100\%    & 50\%     & 0        \\
  \makecell[l]{Need to read cross-$W_K$ and $W_V$ \\ during the generate-phase?} & No & $W_{KV}$ only & $W_K$ and $W_V$ \\
  Support for RoPE?          & Yes      & Yes      & No       \\
  Complexity of cross-phase  & full     & half     & 0        \\ \fline
\end{tabular} \label{tab6} \end{table}

The \textbf{baseline} implementation consists of the following three phases:
\begin{itemize}[topsep=-1pt]
  \item During the \textbf{prompt-phase}, only the encoder is active. All $p$ input-tokens are batched up and processed in parallel. This phase is identical to the prompt-phase of a decoder-only transformer albeit without a causal mask (and is also identical to the entire inference of an encoder-only transformer such as BERT).
  \item During the \textbf{cross-phase}, we take the output of the encoder (which is a $p \times d$ matrix) and precompute the KV projections for the decoder’s cross-attention and store them in cross context memory (aka cross KV-cache).
  \item The \textbf{generate-phase} is similar to the generate-phase of a decoder-only transformer with the following difference: There is a cross-attention block for each layer, which reads keys and values from the cross KV-cache. In addition, the self-attention blocks have their own self KV-cache (which is not the same as the cross KV-cache).
\end{itemize}

\textbf{Option 1} calculates V from K for both self-attention and cross-attention of the decoder stack (note that the encoder stack doesn’t have a KV-cache because it is not autoregressive). This requires reading the cross-$W_{KV}$ parameters from memory during the generate-phase. So if the $W_{KV}$ matrices are larger than the cross V-cache, then calculating V from K for the cross-attention doesn’t make sense for batch size 1 (but for larger batch sizes). For the Whisper models for example, the cross V-cache is always larger than the $W_{KV}$ matrices, because the number of encoder-tokens $p$ is always 1500. And for batch sizes $B$ larger than 1, the reading of the cross-$W_{KV}$ parameters can be amortized among the $B$ inferences.

\begin{figure}[h!] \centering
  \includegraphics[scale=0.71]{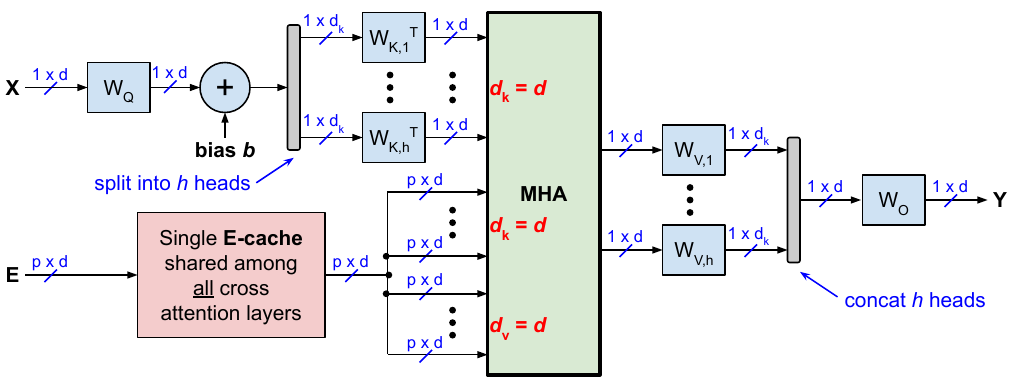}
  \caption{Slim attention with a single E-cache (encoder cache) that is shared among all cross-attention layers for the generate-phase of encoder-decoder transformers.}
\label{fig5} \end{figure}

\textbf{Option 2} efficiently recomputes the cross KV-projections from the encoder output instead of storing them in the cross KV-cache as follows:
\begin{itemize}[topsep=-1pt]
  \item At the end of the prompt-phase, the encoder output is a $p \times d$ matrix, which is then used by all layers of the decoder stack. We call this matrix $E$ \eR{p}{d} the encoder-cache (or E-cache) and we assume that it resides in on-chip SRAM (such as L2 or L3 cache) at the end of the prompt-phase, because it’s usually very small (less than 1 million values for Whisper tiny and base for example).
  \item Recomputing all K-vectors could be done by multiplying $E$ \eR{p}{d} with weight matrix $W_K$ \eR{d}{d}, which requires $2 p d^2$ operations and would be very expensive. A lower complexity option is illustrated in Fig. \ref{fig5}, which is similar to Fig. \ref{fig4}. The main difference is that all cross-attention layers share the same E-cache. And on many machines, this E-cache might fit into on-chip SRAM so that it doesn’t need to be re-read for each layer during the generate-phase.
  \item As with Fig. \ref{fig4} in the previous section, Fig. \ref{fig5} doesn’t support RoPE, but other (and potentially better) relative PE schemes such as RPE and FIRE.
  \item This scheme doesn’t calculate V from K and therefore is not limited to MHA only or to projection matrices that can be inverted.
  \item Similar to option 1 (and unlike the baseline), the cross-$W_V$ and $W_K$ parameters need to be read from memory for each generated token during the generate-phase. Therefore for batch size 1, this scheme only makes sense if the KV-cache is larger than the number of cross-$W_V$ and $W_K$ parameters, which is the case for all Whispers models (because they use $p = 1500$ input-tokens, and $d$ of the largest Whisper model is smaller than 1500). And for batch sizes larger than 1, this option usually makes sense because the parameter reads are amortized among all the inferences of the batch.
\end{itemize}

\textbf{Time-to-first-token (TTFT)}: Options 1 and 2 speed up or eliminate the cross-phase. Specifically, option 1 speeds up the cross-phase by 2x. And option 2 completely eliminates the cross-phase. This speeds up the time-to-first-token latency (TTFT). And in certain cases, the overall compute complexity can be reduced compared to the baseline. For cases where the number of encoder-tokens is larger than the decoder-tokens (such as for Whisper or for text summarization) and for large $d$ dimensions, these options can reduce the overall complexity.

Table \ref{tab7} lists the cache size savings and speedups during the generate-phase for all 5 Whisper models, assuming a fixed encoder context length of $p = 1500$, a decoder output context length of 448, and a vocab\_size of 51,865. Option 2 reduces the cache sizes by 8.7x, assuming that the entire encoder output (matrix E) is kept in on-chip SRAM (the E-cache), which speeds up the generate-phase by over 5x for a memory bound system with batch size 64. Note that the speedups listed in the table are in addition to speeding up the cross-phase (i.e. 2x cross-phase speedup for option 1, and entire elimination of the cross-phase for option 2).

\begin{table} \centering
  \caption{Cache size savings and speedups of generate phase for Whisper models}
  \begin{tabular}{lcccccl} \fline
  & \mc{5}{\thead{Whisper models}} & \\
  & \thead{tiny} & \thead{base} & \thead{small} & \thead{medium} & \thead{large} & \thead[l]{Notes} \\ \hline
  Params     & 38M   & 73M   & 242M  & 764M   & 1.5B  & number of parameters                   \\
  $l$        & 4     & 6     & 12    & 24     & 32    & number of layers                       \\
  $d$        & 384   & 512   & 768   & 1,024  & 1,280 & embedding dimension                    \\
  $d_{\text{ffn}}$ & 1,536 & 2,048 & 3,072 & 4,096  & 5,120 & hidden dimension of FFN            \\ \hline
  \multicolumn{7}{l}{\thead[l]{Cache sizes (in M):}}                                           \\ \hline
  Encoder E-cache  & 0.6   & 0.8  & 1.2  & 1.5 & 1.9 & $1500 \cdot d$                          \\
  Cross KV-cache   & 4.6   & 9.2  & 27.6 & 73.7 & 122.9 & $2 \cdot 1500 \cdot d \cdot l$       \\
  Self KV-cache    & 1.4   & 2.8  & 8.3  & 22.0 & 36.7  & $2 \cdot 448 \cdot d \cdot l$        \\
  Baseline cache   & 6.0   & 12.0 & 35.9 & 95.7 & 159.6 & cross KV + self KV                   \\
  Option 1 cache   & 3.0   & 6.0  & 18.0 & 47.9 & 79.8 & half of baseline                      \\
  Option 2 cache   & 0.7   & 1.4  & 4.1  & 11.0 & 18.4 & no cross KV + half of self KV         \\
  Option 2 savings & \textbf{8.7x}  & \textbf{8.7x} & \textbf{8.7}x & \textbf{8.7x} & \textbf{8.7x} & cache savings vs. baseline \\ \hline
  \multicolumn{7}{l}{\thead[l]{Number of parameters (in M) for generate-phase:}}               \\ \hline
  Baseline params  & 28.2 & 48.6 & 138.9 & 405.4 & 800.4 & $d \cdot \text{vocab} + l \cdot (6d^2 + 2d \cdot d_{\text{ffn}})$ \\
  Option 1 params  & 28.8 & 50.1 & 146.0 & 430.6 & 852.8 & baseline + cross K ($d^2 \cdot l$)   \\
  Option 2 params  & 29.4 & 51.7 & 153.1 & 455.8 & 905.2 & baseline + cross KV ($2d^2 \cdot l$) \\ \hline
  \multicolumn{7}{l}{\thead[l]{Memory reads (in M) per token for batch size 1:}}               \\ \hline
  Baseline         & 34.2  & 60.5  & 174.8 & 501.2 & 960.0 & baseline cache + baseline params  \\
  Option 1         & 31.8  & 56.1  & 164.0 & 478.5 & 932.6 & option 1 cache + option 1 params  \\
  Option 2         & 30.0  & 53.1  & 157.2 & 466.8 & 923.6 & option 2 cache + option 2 params  \\
  Option 1 speedup & 1.08x & 1.08x & 1.07x & 1.05x & 1.03x & speedup vs. baseline              \\
  Option 2 speedup & 1.14x & 1.14x & 1.11x & 1.07x & 1.04x & speedup vs. baseline              \\ \hline
  \multicolumn{7}{l}{\thead[l]{Memory reads (in M) per token for batch size 64:}}              \\ \hline
  Baseline         & 6.4  & 12.7 & 38.1 & 102.1 & 172.1 & baseline cache + $1/64 \cdot$ params \\
  Option 1         & 3.4  & 6.8  & 20.2 & 54.6  & 93.1  & option 1 cache + $1/64 \cdot$ params \\
  Option 2         & 1.1  & 2.2  & 6.5  & 18.1  & 32.5  & option 2 cache + $1/64 \cdot$ params \\
  Option 1 speedup & 1.9x & 1.9x & 1.9x & 1.9x  & 1.8x  & speedup vs. baseline                 \\
  Option 2 speedup & \textbf{5.6x} & \textbf{5.8x} & \textbf{5.8x} & \textbf{5.6x} & \textbf{5.3x} & speedup vs. baseline \\ \fline
\end{tabular} \label{tab7} \end{table}

\section{Experiments} \label{sec:experiments}
This section compares the K-cache scheme of equation (\ref{eq6}) with the V-cache scheme of appendix \ref{sec:vcache} for the decoder cross-attention of Whisper tiny, small, medium, and large-v3 \citep{whisper}. Whisper's V projection has a bias $b_V$ and its K projection does not. So the K-cache scheme uses ${V = K W_{KV} + b_V}$ and the V-cache scheme uses ${K = (V - b_V) W_{VK}}$. Alternatively, $b_V$ can be removed from both schemes as in appendix \ref{sec:bias}. The matrices $W_{KV}$ and $W_{VK}$ are precomputed in float64 and then converted to the precision of the model.

Table \ref{tab:cond} lists the largest condition number $\kappa$ of $W_K$ and of $W_V$ (the ratio of the largest to the smallest singular value) and the largest $\lvert\det W_K\rvert$, which is below $10^{-230}$ in every model, including tiny, whose $W_K$ is the best conditioned. See \texttt{slimAttn\_whisper.py} in \citep{tricks} for the per-layer condition numbers.

\begin{table}[h!] \centering
  \caption{Condition numbers of $W_K$ and $W_V$ and determinant of $W_K$ in the cross-attention of the Whisper decoder}
  \begin{tabular}{lcccc} \fline
                                          & tiny             & small            & medium           & large-v3         \\ \hline
  Largest $\kappa(W_K)$                   & $3.7 \cdot 10^4$ & $3.3 \cdot 10^7$ & $1.1 \cdot 10^7$ & $5.2 \cdot 10^9$ \\
  Largest $\kappa(W_V)$                   & $5.4 \cdot 10^3$ & $8.0 \cdot 10^4$ & $6.2 \cdot 10^5$ & $4.5 \cdot 10^6$ \\
  Largest $\lvert\det W_K\rvert$          & $< 10^{-231}$    & $< 10^{-473}$    & $< 10^{-760}$    & $< 10^{-744}$    \\
  Layers with $\kappa(W_V) < \kappa(W_K)$ & 3 of 4           & 9 of 12          & 21 of 24         & 31 of 32         \\ \fline
\end{tabular} \label{tab:cond} \end{table}

Table \ref{tab:logits} lists the relative logit error, which is the largest absolute difference between the logits of the model with slim attention and the logits of the baseline, divided by the largest absolute logit of the baseline, with both maxima taken over all utterances, positions, and vocabulary entries. The encoder input is 4 dev-clean utterances from LibriSpeech \citep{librispeech}, one from each of 4 speakers, and the decoder input is Whisper's four start tokens followed by 16 tokens that the baseline generates greedily for English transcription, with end-of-text suppressed. The float64 row runs the decoder and its baseline in float64 on the same encoder output, converted to float64, and the same decoder input. The per-layer choice uses the K-cache in layers with $\kappa(W_K) < \kappa(W_V)$ and the V-cache in all other layers. For reference, replacing each cross-attention with an exact float32 implementation that uses the true K and V gives a relative logit error of about $10^{-6}$ on the same inputs.

\begin{table}[h!] \centering
  \caption{Relative logit error of slim attention in the decoder cross-attention on LibriSpeech, in float32 unless noted}
  \begin{tabular}{lcccc} \fline
                               & tiny                 & small                & medium               & large-v3             \\ \hline
  K-cache in every layer       & $4.98 \cdot 10^{-4}$  & $2.24 \cdot 10^{-2}$  & $4.43 \cdot 10^{-3}$  & $1.09$                \\
  V-cache in every layer       & $9.85 \cdot 10^{-6}$  & $5.14 \cdot 10^{-5}$  & $7.01 \cdot 10^{-5}$  & $1.54 \cdot 10^{-3}$  \\
  Per-layer choice             & $7.62 \cdot 10^{-5}$  & $9.22 \cdot 10^{-5}$  & $1.62 \cdot 10^{-4}$  & $2.56 \cdot 10^{-3}$  \\
  Per-layer choice, float64    & $1.95 \cdot 10^{-13}$ & $1.99 \cdot 10^{-13}$ & $1.40 \cdot 10^{-12}$ & $1.39 \cdot 10^{-11}$ \\ \fline
\end{tabular} \label{tab:logits} \end{table}

In Table \ref{tab:logits}, the error of the V-cache in every layer is about 50x (tiny) to about 700x (large-v3) smaller than the error of the K-cache in every layer. For large-v3, the K-cache in every layer has an error of 1.09, so its largest absolute logit difference exceeds the largest absolute logit of the baseline. The per-layer choice differs from the V-cache in every layer only in the 1 to 3 layers that it assigns to the K-cache, and those layers make its error 1.7x to 7.7x larger. On 4 utterances from 4 other speakers, the K-cache in every layer again has the largest error in every model, and the per-layer choice has a larger error than the V-cache in every layer in three of the four models, all but small. So a smaller condition number does not guarantee a smaller error: choosing per layer by condition number gave a larger error than the V-cache in every layer in 7 of these 8 cases. The float64 row is more than five orders of magnitude below every float32 entry, and in float64 the K-cache and the V-cache in every layer also stay below $10^{-8}$, so the float32 errors vanish with more precision.

\section{Conclusion}
Slim attention offers a simple trick for halving the context memory of existing MHA transformer models without sacrificing accuracy. Slim attention is a post-training, exact implementation of existing models, so it doesn't require any fine-tuning or training from scratch.

Future work includes integrating slim attention into popular frameworks such as HuggingFace Transformers \citep{HFtransformers}, whisper.cpp \citep{whisper-cpp}, llama.cpp \citep{llama-cpp}, vLLM \citep{vLLM}, llamafile \citep{llamafile}, LM Studio \citep{lmstudio}, Ollama \citep{ollama}, SGLang \citep{sglang}, and combining it with existing context memory management schemes such as PagedAttention \citep{pagedAttn} and other compression schemes such as Dynamic Memory Compression DMC \citep{DMC} and VL-cache \citep{VL-cache}.

Please also see our forthcoming paper about matrix-shrink \citep{matShrink}, which reduces the sizes of attention matrices and proposes a simplification of DeepSeek's MLA scheme.

\section*{Acknowledgments}
We would like to thank \href{https://scholar.google.com/citations?user=KEhvGNMAAAAJ&hl=en}{Dirk Groeneveld (Ai2)} and \href{https://scholar.google.com/citations?user=Be2fl8sAAAAJ&hl=en}{Zhiqiang Xie (Stanford, SGLang)} for helpful discussions on this work. We are particularly grateful to \href{https://scholar.google.com/citations?user=wsGvgA8AAAAJ&hl=en}{Noam Shazeer} for his positive feedback on slim attention. And special thanks to Imo Udom, Mitchell Baker, and Mozilla for supporting this work.

\appendix

\section{Matrix inversion experiments}
The code listed below demonstrates that V can be accurately computed from K for each layer of the exemplary model SmolLM2-1.7B from HuggingFace \citep{smollm}. The last three lines of the code listing compute $W_{KV} = W_K^{-1} W_V$ and then compare the original $W_V$ against its reconstruction $W_K W_{KV}$ by using the NumPy function allclose(). Running below code in a free Jupyter notebook on the default CPU takes about 3 minutes (including downloading the model) and takes 1.7GB of system DRAM.

\begin{python}
import transformer_tricks as tt
import numpy as np
import torch
from transformers import AutoConfig

repo = 'HuggingFaceTB/SmolLM2-1.7B'
param = tt.get_param(repo)
config = AutoConfig.from_pretrained(repo)

# check if we can accurately compute V from K for each layer
for layer in range(config.num_hidden_layers):
  # convert to float64 for better accuracy of matrix inversion
  # note that all weight-matrices are transposed in tensorfile
  Wk = param[tt.weight('K', layer)].to(torch.float64).numpy().T
  Wv = param[tt.weight('V', layer)].to(torch.float64).numpy().T
  Wkv = np.linalg.inv(Wk) @ Wv
  # check if Wk @ Wkv is close to Wv
  print(layer, ':', np.allclose(Wk @ Wkv, Wv))
\end{python}

\section{MHA complexity}
This section derives the number of OPs (two-operand operations) for the column labeled ``MHA complexity'' of Table \ref{tab2}, which is the MHA-complexity per token during the generate-phase. Recall that multiplying two matrices (MatMul) with dimensions $m \times n$ and $n \times p$ entails computing $m p$ dot-products of length $n$, where each dot-product takes $n$ MULs (two-operand multiply operations) and $n-1$ ADDs (two-operand add operations). In total, the entire MatMul requires $mnp$ MULs and $mp(n-1)$ ADDs, so $mp(2n-1)$ OPs, which is approximately $2mnp$ OPs.
\begin{table}[h!] \centering
  \caption{OPs per attention head for MHA during the generate phase}
  \begin{tabular}{lccc} \fline
                         & Term & Dimensions of MatMul   & OPs per attention-head                    \\ \hline
  Softmax argument       & $Q_i K_i^\top$                & $(1 \times d_k)(d_k \times n)$   & $2nd_k$ \\
  Weighted sum of $V_i$  & $\text{softmax}() \cdot V_i$  & $(1 \times n)(n \times d_k)$    & $2nd_k$ \\
  Weighted sum of $K$    & $\text{softmax}() \cdot K$    & $(1 \times n)(n \times d)$      & $2nd$   \\ \fline
\end{tabular} \label{tab8} \end{table}

Table \ref{tab8} specifies the number of OPs per attention-head for computing the softmax arguments and the weighted sums. This is for the generate-phase where all MatMuls are actually vector-matrix products (instead of matrix-matrix products).

Table \ref{tab9} shows the total number of MHA OPs across all attention-heads for the three cases of Table \ref{tab2} (i.e. for Vanilla, unoptimized Fig. \ref{fig3}(b), and optimized Fig. \ref{fig3}(c)). Note that $d_k = d / h$.
\begin{table}[h!] \centering
  \caption{Total OPs for MHA during the generate phase}
  \begin{tabular}{lccc} \fline
                                  & \mc{2}{OPs per attention-head}   &                            \\
                                  & softmax argument & weighted sum  & Total OPs                  \\ \hline
  Vanilla, see Fig. \ref{fig2}(b) & $2nd_k$          & $2nd_k$       & $h(2nd_k + 2nd_k) = 4nd$    \\
  Unoptimized, Fig. \ref{fig3}(b) & $2nd_k$          & $2nd_k$       & $4nd$                       \\
  Optimized, Fig. \ref{fig3}(c)   & $2nd_k$          & $2nd$         & $h(2nd_k + 2nd) = 2nd(h+1)$ \\ \fline
\end{tabular} \label{tab9} \end{table}

\section{Alternative scheme: \emph{all you need is V-cache}} \label{sec:vcache}
\begin{figure}[h!] \centering
  \includegraphics[scale=0.9]{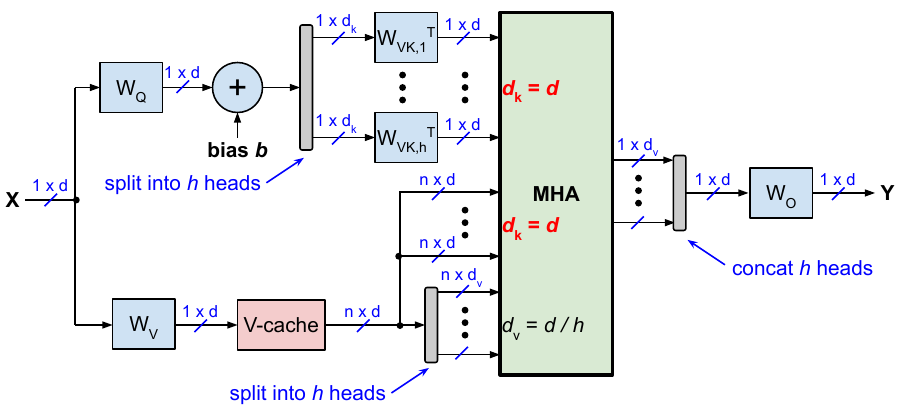}
  \caption{Alternative scheme during generate-phase}
\label{fig6} \end{figure}

Instead of calculating V from K, it's also possible to calculate K from V and thereby eliminate the K-cache (instead of the V-cache). This alternative scheme is illustrated in Fig. \ref{fig6}, where $W_{VK} = W_V^{-1} W_K$. However, this scheme does not support RoPE.

\section{Taking advantage of softmax sparsities}
In this section we describe how we can take advantage of softmax sparsities (i.e. sparsities in the attention scores) to reduce the computational complexity of the attention blocks. In some applications, many attention scores are 0 or close to zero. For those attention scores (i.e. attention scores smaller than a threshold), we can simply skip the corresponding V-vector, i.e. we don’t have to add those skipped vectors to the weighted sum of V-vectors. This reduces the complexity of calculating the weighted sum of V-vectors. For example, for a sparsity factor $S = 0.8$ (i.e. 80\% of scores are 0), the complexity is reduced by factor $\frac{1}{1 - S} = 5$.

By the way, taking advantage of softmax sparsities is also possible for systems with KV-cache where V is not computed from K. In this case, skipping V-vectors with zero scores means that we don’t have to read those V-vectors from the KV-cache, which speeds up the autoregressive generate-phase for memory bound systems. However, this will never speed it up more than slim attention's removal of the entire V-cache. Furthermore, for MQA and GQA, each V-vector is shared among multiple (e.g. 4 or more) queries so we can only skip reading a V-vector from memory if all 4 (or more) attention scores are zero for this shared V-vector, which reduces the savings significantly. For example, if the V-vectors are shared among 4 queries and the attention scores have sparsity $S = 0.8$, then the probability of all four queries being 0 is only $S^4 = 0.41$, so we can only skip 41\% of the V-vectors.

\section{Support for RoPE}
Many transformers nowadays use RoPE (rotary positional embedding) \citep{RoPE}, which applies positional encoding to the Q and K projections, but not the V projections. In general, RoPE can be applied to the K projections either before storing them in K-cache or after reading them from K-cache. The former is preferred because of lower computational complexity during the generate-phase (so that each K-vector is RoPE’d only once instead of multiple times). However, if the RoPE’d keys are stored in K-cache, then we first need to un-RoPE them before we can compute V from K. The following details two options to support RoPE.

\textbf{Option 1} is for the case where we don’t take advantage of softmax sparsities. In this case, we apply RoPE to the K-vectors after reading them from K-cache during the generate-phase. That way we can use the raw K-vectors for computing V from K.

\textbf{Option 2} is for the case where we take advantage of softmax sparsities as detailed in the previous section. In this case, RoPE is applied to the K-vectors before writing them into the K-cache. And when they are read from K-cache during the generate-phase, then we have to revert (or undo) the RoPE-encoding before we can use the K-vectors to compute the V-vectors (i.e. multiplying the K-vectors with the attention scores). However, we only need to do this for a portion of the K-vectors, depending on the sparsity factor $S$. For example, for $S = 0.8$, we only need to revert the RoPE-encoding for 20\% of the K-vectors. The RoPE encoding can be reverted (aka RoPE-decoding) by simply performing a rotation in the opposite direction by the same amount as shown below for the 2D case.

\textbf{RoPE encoding}:
\begin{align*}
  y_1 &=  x_1 \cos{m \theta} + x_2 \sin{m \theta} \\
  y_2 &= -x_1 \sin{m \theta} + x_2 \cos{m \theta}
\end{align*}

\textbf{RoPE decoding}:
\begin{align*}
  x_1 &= y_1 \cos{m \theta} - y_2 \sin{m \theta} \\
  x_2 &= y_1 \sin{m \theta} + y_2 \cos{m \theta}
\end{align*}
Note that the RoPE decoding uses the same trigonometric coefficients (such as $\cos{m \theta}$) as the RoPE encoding. Therefore, we only need one look-up table that can be used for both RoPE encoding and decoding.

\section{Support for bias} \label{sec:bias}
Since PaLM’s removal of bias terms from all its projection layers \citep{PaLM}, most transformer models nowadays do the same. However, some models are still using biases today (especially older models that are still relevant today such as Whisper). In this section, we briefly discuss how projection layers with bias can be supported. We show how the biases of two of the four attention projection layers can be eliminated in a mathematically equivalent way.

\textbf{Bias removal for V projections}: This bias can be combined with the bias of the output projection layer as follows. Recall that all value vectors $v_i$ plus their constant bias $b$ are multiplied by the attention scores $s_i$ (i.e. the softmax outputs) and summed up, such as
\begin{equation*}
  \sum_{i=1}^n s_i (v_i + b) = \sum_{i=1}^n s_i v_i + \sum_{i=1}^n s_i b = \sum_{i=1}^n s_i v_i + b
\end{equation*}

The last equal sign holds because the sum over all attention-scores $s_i$ is always 1 as per softmax definition (because the softmax generates a probability distribution that always adds up to 1). We can now merge the bias $b$ with bias $c$ of the preceding output projection layer (O) as follows: $y = (x + b) W_O + c = x W_O + (b W_O + c) = x W_O + c^\ast$, with the new bias $c^\ast = b W_O + c$. This new bias-vector $c^\ast$ can be computed offline, before inference time. Or simply remove the V-bias already during training.

\textbf{Bias removal for K projections}: The bias of the K projection cancels out due to the constant invariance of the softmax function. For example, say we have 2-dimensional heads, then the dot-product $p$ between query-vector $q = (q_1 + b_1, q_2 + b_2)$ with bias $b$ and key-vector $k = (k_1 + c_1, k_2 + c_2)$ with bias $c$ is as follows:
\begin{align*}
  p &= (q_1 + b_1)(k_1 + c_1) + (q_2 + b_2)(k_2 + c_2) \\
    &= [q_1k_1 + q_2k_2] + [q_1c_1 + q_2c_2] + [b_1k_1 + b_2k_2] + [b_1c_1 + b_2c_2] \\
    &= q_1k_1 + q_2k_2 + f(q) + b_1k_1 + b_2k_2 + \text{constant},
\end{align*}
where $f(q) = q_1c_1 + q_2c_2$ is a function of the query-vector only; and “constant” is a constant that only depends on the two biases $b$ and $c$. Now recall that the softmax function doesn’t change if a constant is added to all its arguments. Because all arguments of the attention softmax use the same single query-vector $q$, $f(q)$ is the same for all arguments and is therefore constant and can be removed from all softmax arguments. As a result, we can remove the entire bias-vector $c$ from the keys. But we still need the bias-vector $b$ for the queries. However, this assumes that there is no RoPE applied between the projections and the dot-product calculation, which is fortunately the case for Whisper for example.

\section{Slim Attention for GQA (such as Gemma2-9B)}
Slim attention is not limited to MHA only. In general, slim attention reduces the KV-cache size whenever the width of the KV-cache ($d_\text{cache}$) is larger than $d_\text{model}$, where $d_\text{cache} = h_{KV} (d_K + d_V)$, where $h_{KV}$ is the number of KV-heads. In this case, slim attention can compress the KV-cache size by the compression factor $c = d_\text{cache} / d_\text{model}$. Usually, this compression factor $c$ is 2. In some cases, $c$ is larger than 2, for example CodeGemma-7B has $c = 2*4096 / 3072 = 2.67$.

For Gemma2-9B and PaliGemma2-10B, which utilize GQA instead of MHA, $c = 2*2048/3584 = 1.14$. In this case where the compression factor $c$ is smaller than 2 and larger than 1, the application of slim attention is not straight forward. The figure below shows how we can implement slim attention for this case, where $e = h_{KV} d_K$ is the projection dimension of the $d \times e$ weight matrices and $d/2 < e < d$. For Gemma2-9B, $e = 2048$ and $d = 3584$. Note that $W_K^\ast$ has $d^2$ parameters, while the original $W_K$ has only $d e$ parameters, which is a disadvantage.
\begin{figure}[h!] \centering
  \includegraphics[scale=1.0]{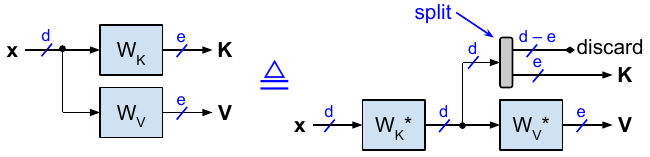}
  \caption{Computing V from K for the case where the KV-cache width $2e$ is larger than $d$}
\label{fig7} \end{figure}

\section{Further optimization for encoder-decoder transformers}
This section details a further optimization, which is a hybrid of options 1 and 2 of section \ref{sec:enc-dec}, where one layer of the decoder stack uses option 1 and the other layers use a modified version of option 2 as follows: Say the first layer implements option 1, which entails calculating and storing the cross K-cache for the first layer, i.e. $K = E W_K$. The trick is now to use this K-cache instead of the E-cache for all the other layers and then we can calculate E from K as $E = K W_{K\_layer1}^{-1}$. So the other layers will now use K instead of E, which means that they need to use modified versions of their $W_K$ and $W_V$ weight matrices, which are offline computed as $W_{V\_new} = W_{K\_layer1}^{-1} W_{V\_old}$ and $W_{K\_new} = W_{K\_layer1}^{-1} W_{K\_old}$. The savings of this hybrid versus option 2 are not huge: It only speeds up the first layer because we only need to read one of the two cross weight matrices for the first layer for each generated token (i.e. we only need to read $W_{KV\_layer1}$ instead of reading both $W_{K\_layer1}$ and $W_{V\_layer1}$). So this hybrid makes only sense for shallow models, e.g. Whisper tiny which only has 4 layers.

\section{More MHA models}
Table \ref{tab10} lists additional transformer models with MHA, similar to Table \ref{tab1}. See HuggingFace for more details on these models.

\def\METAGENE   {\href{https://huggingface.co/metagene-ai/METAGENE-1}                   {METAGENE-1 }}
\def\OLMoE      {\href{https://huggingface.co/allenai/OLMoE-1B-7B-0125}                 {OLMoE-1B-7B-0125 }}
\def\llmcomp    {\href{https://huggingface.co/facebook/llm-compiler-13b}                {llm-compiler-13B }}
\def\drama      {\href{https://huggingface.co/facebook/drama-base}                      {DRAMA-base }}
\def\chameleon  {\href{https://huggingface.co/facebook/chameleon-7b (Meta)}             {Chameleon-7B }}
\def\DeepSeekvl {\href{https://huggingface.co/deepseek-ai/deepseek-vl2-tiny}            {DeepSeek-vl2-tiny }}
\def\Crisper    {\href{https://huggingface.co/nyrahealth/CrisperWhisper}                {CrisperWhisper }}
\def\Moonshine  {\href{https://huggingface.co/UsefulSensors/moonshine-base}             {Moonshine-base }}
\def\Moondream  {\href{https://huggingface.co/vikhyatk/moondream2}                      {Moondream-2 }}
\def\StableCode {\href{https://huggingface.co/stabilityai/stable-code-3b}               {Stable-Code-3B }}
\def\StableLM   {\href{https://huggingface.co/stabilityai/stablelm-2-1_6b}              {Stable-LM2-1.6B }}
\def\OpenMath   {\href{https://huggingface.co/nvidia/OpenMath-CodeLlama-13b-Python-hf}  {OpenMath-CodeLlama-13B }}
\def\MAIRA      {\href{https://huggingface.co/microsoft/maira-2}                        {MAIRA-2 }}
\def\Rocket     {\href{https://huggingface.co/pansophic/rocket-3B}                      {Rocket-3B }}
\def\OpenLlama  {\href{https://huggingface.co/openlm-research/open_llama_13b}           {OpenLLaMA-13B }}
\def\TimesFM    {\href{https://huggingface.co/google/timesfm-1.0-200m}                  {TimesFM-1.0-200M }}
\def\MetricX    {\href{https://huggingface.co/google/metricx-24-hybrid-xxl-v2p6}        {MetricX-24-hybrid-XXL }}
\def\BioMedLM   {\href{https://huggingface.co/stanford-crfm/BioMedLM}                   {BioMedLM }}
\def\MiniCPM    {\href{https://huggingface.co/openbmb/MiniCPM-V-2}                      {MiniCPM-V2 }}
\def\UltraLM    {\href{https://huggingface.co/openbmb/UltraLM-65b}                      {UltraLM-65B }}
\def\TogetherAI {\href{https://huggingface.co/togethercomputer/LLaMA-2-7B-32K}          {LLaMA-2-7B-32K }}
\def\Dolly      {\href{https://huggingface.co/databricks/dolly-v2-12b}                  {Dolly-v2-12B }}
\def\MPTseven   {\href{https://huggingface.co/mosaicml/mpt-7b-8k}                       {MPT-7B-8K }}
\def\MPTthirty  {\href{https://huggingface.co/mosaicml/mpt-30b}                         {MPT-30B }}
\def\BLOOM      {\href{https://huggingface.co/bigscience/bloom}                         {BLOOM }}

\begin{table}[h!] \centering
  \caption{Additional transformer models with MHA}
  \begin{tabular}{lllccc} \fline
  \thead[l]{Year} & \thead[l]{Publisher} & \thead[l]{Model} & \thead{Params} & \thead{$d$} & \thead{layers} \\ \hline
  2025 & Metagene        & \METAGENE \citep{metagene}       & 6.5B           & 4096        & 32             \\
  2025 & Ai2             & \OLMoE \citep{olmoe}             & 6.9B           & 2048        & 16             \\
  2025 & Meta            & \drama \citep{drama}             & 212M           & 768         & 12             \\
  2024 & Meta            & \chameleon \citep{chameleon}     & 7B             & 4096        & 32             \\
  2024 & Meta            & \llmcomp \citep{llmCompiler}     & 13B            & 5120        & 40             \\
  2024 & DeepSeek        & \DeepSeekvl                      & 3.4B           & 1280        & 12             \\
  2024 & nyra health     & \Crisper \citep{crisper}         & 1.6B           & 1280        & 32             \\
  2024 & Useful Sensors  & \Moonshine \citep{moonshine}     & 61M            & 416         & 8              \\
  2024 & Moondream       & \Moondream                       & 1.9B           & 2048        & 24             \\
  2024 & Stability AI    & \StableCode \citep{stableCode}   & 2.8B           & 2560        & 32             \\
  2024 & Stability AI    & \StableLM \citep{stableLM}       & 1.6B           & 2048        & 24             \\
  2024 & NVIDIA          & \OpenMath \citep{openMath}       & 13B            & 5120        & 40             \\
  2024 & Microsoft       & \MAIRA \citep{maira2}            & 6.9B           & 4096        & 32             \\
  2024 & Pansophic       & \Rocket                          & 2.8B           & 2560        & 32             \\
  2024 & OpenLM Research & \OpenLlama                       & 13B            & 5120        & 40             \\
  2024 & Google          & \TimesFM \citep{timesFM}         & 200M           & 1280        & 20             \\
  2024 & Google          & \MetricX \citep{metricX}         & 13B            & 4096        & 24             \\
  2024 & Stanford        & \BioMedLM \citep{biomedLM}       & 2.7B           & 2560        & 32             \\
  2024 & OpenBMB         & \MiniCPM \citep{miniCPMv2}       & 3.4B           & 2304        & 40             \\
  2023 & OpenBMB         & \UltraLM                         & 65B            & 8192        & 80             \\
  2023 & Together AI     & \TogetherAI                      & 7B             & 4096        & 32             \\
  2023 & Databricks      & \Dolly                           & 12B            & 5120        & 36             \\
  2023 & Mosaic ML       & \MPTseven                        & 7B             & 4096        & 32             \\
  2023 & Mosaic ML       & \MPTthirty                       & 30B            & 7168        & 48             \\
  2022 & BigScience      & \BLOOM \citep{bloom}             & 176B           & 14,336      & 70             \\ \fline
\end{tabular} \label{tab10} \end{table}


\bibliographystyle{unsrtnat}
\bibliography{references}

\end{document}